# The Synthinel-1 dataset: a collection of high resolution synthetic overhead imagery for building segmentation

*Fanjie Kong[1], Bohao Huang[1], Kyle Bradbury[2], Jordan M. Malof[1]*

[1]Department of Electrical & Computer Engineering, Duke University, Durham, NC 27708
[2]Energy Initiative, Duke University, Durham, NC 27708

## Abstract

*Recently deep learning – namely convolutional neural networks (CNNs) – have yielded impressive performance for the task of building segmentation on large overhead (e.g., satellite) imagery benchmarks. However, these benchmark datasets only capture a small fraction of the variability present in real-world overhead imagery, limiting the ability to properly train, or evaluate, models for real-world application. Unfortunately, developing a dataset that captures even a small fraction of real-world variability is typically infeasible due to the cost of imagery, and manual pixel-wise labeling of the imagery. In this work we develop an approach to rapidly and cheaply generate large and diverse virtual environments from which we can capture synthetic overhead imagery for training segmentation CNNs. Using this approach, generate and publicly-release a collection of synthetic overhead imagery – termed Synthinel-1 with full pixel-wise building labels. We use several benchmark dataset to demonstrate that Synthinel-1 is consistently beneficial when used to augment real-world training imagery, especially when CNNs are tested on novel geographic locations or conditions.*

## 1. Introduction

Building footprint segmentation in overhead imagery (e.g., satellite images, aerial photography) is a challenging problem that has been extensively investigated within the computer vision community [1]–[3]. Recently, convolutional neural networks (CNNs) have led to substantial performance improvements over previous segmentation methods, and CNNs now dominate benchmark problems [4]–[6]. CNNs are high-capacity non-linear models that must be trained to perform segmentation using large quantities of overhead imagery in which the building footprints have been annotated (e.g., with polygons). Therefore, a crucial contributor to the recent success of CNNs for building segmentation has been the development of large publicly-available benchmark datasets of *hand-labeled* overhead imagery. Recent datasets such as Inria [7], DSTL [6], and DeepGlobe [5] are unprecedented in their size and geographic coverage.

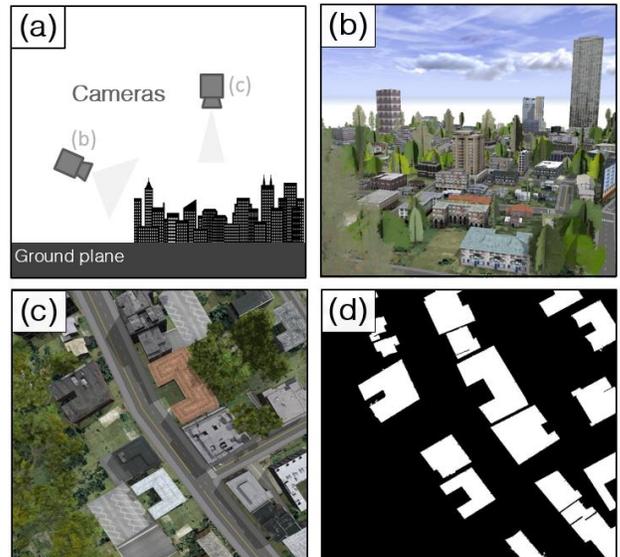

Fig. 1. (a) Illustration of a virtual city and two perspectives of a virtual camera, set by the designer. The corresponding images for each camera are shown in panels (b) and (c). The camera settings in (c) are designed to create images that mimic overhead imagery. We changed the focal length of the camera (illustrated in (a)) so that the imagery mimics the perspective of a camera located at a much higher altitude than can be achieved with our virtual camera. In (d) we show the corresponding ground truth labels extracted for the image in (c), which are readily available because we designed all of the content in the virtual world.

Despite their unprecedented scope, modern benchmark datasets still encompass relatively little of the variability present in real-world overhead imagery. The visual characteristics of overhead imagery vary tremendously, due to numerous factors: imaging conditions (e.g., camera NADIR, spatial resolution), environmental conditions (e.g., weather and atmospheric conditions, time-of-day, season), and geographic location (e.g., regional building styles vary across the globe). Recent benchmark datasets, however, all encompass just a few geographic locations, and each location is imaged under relatively uniform conditions.

Due to these limitations, it is unclear whether segmentation models trained and evaluated on recent benchmark datasets generalize well to novel overhead imagery, arguably a more realistic and practical scenario – recent work [7], [8] indicates that they do not. This result is further corroborated by the work presented here. This represents an important limitation of existing models.

Unfortunately however, collecting a representative set of labeled overhead imagery, reflecting the variability of real-world imagery, is completely infeasible. This would require collecting imagery from locations across the globe, and doing so several times under different imaging conditions. As we discuss in Section 3.2, acquiring even a relatively small imagery dataset, and annotating, is costly.

## 1.1. Synthetic overhead imagery for building labeling

In this work we explore the use of *synthetic* overhead imagery to overcome the limitations of real-world imagery. Here "synthetic imagery" refers to imagery that has been captured from a simulated camera operating over a virtual world, as illustrated in Fig. 1. In a virtual world, a designer can specify the locations and visual characteristics of scene content, as well as the camera location and its characteristics. As a result, the designer can collect large quantities of diverse imagery at little cost. Furthermore, there is no need for annotation since the locations of all objects is known by design.

Recently computer vision researchers have found tremendous success using "synthetic" imagery for training recognition models in several application areas [9], [10], such as object recognition in street [9]–[12] and indoor [13], [14] scenes. However, practical generation of overhead synthetic imagery presents several unique challenges compared to existing use-cases of synthetic imagery; to our knowledge the challenges of overhead imagery remain unexplored in any previous research literature.

Existing uses of synthetic imagery require (geographically) small virtual worlds, with a focus on high-fidelity visual features, layouts, and randomness of *small-scale* objects (e.g., people, vegetation, road signs, furniture, etc.[9], [14]). By contrast, even a small quantity of synthetic overhead imagery (e.g., 10-20 km$^2$) requires a virtual world of corresponding large size. Similarly, overhead synthetic imagery requires visual fidelity at much larger scales (e.g., a few meters), as well as appropriate sources of randomness for such scales (e.g., shape, colors, sizes, and layouts of roads, buildings, and landscapes). Existing tools cannot *efficiently* generate synthetic imagery at the scales required for overhead synthetic imagery.

Tools are needed that can quickly generate large-scale virtual worlds with realistic (large-scale) variations, while maintaining realistic (large-scale) layouts and structure, and providing the designer with high-level controls over important characteristics. Furthermore, while it seems likely that synthetic imagery should *ultimately* be beneficial for training DL models, the process for generating such a dataset, or the properties a virtual world should possess (e.g., layouts, textures, colors), is far from obvious. Object recognition (buildings, or other objects) in overhead imagery relies on unique visual cues and considerations compared to existing work with synthetic imagery on indoor/street scenes.

## 1.2. Contributions of this work

In this work we explore the use of a widely-available software, CityEngine, as a tool for rapidly generating large-scale virtual worlds; it possesses many of the aforementioned capabilities. We develop additional software to rapidly extract overhead imagery at a desired resolution from these virtual worlds. Using these approaches we generated a collection of synthetic imagery, termed Synthinel-1 (inspired by the Sentinel satellites). We demonstrate that Synthinel-1 is beneficial for training modern deep learning models for building segmentation using several recent benchmark datasets and deep learning models. We began with building segmentation due to its popularity, but these approaches can easily be extended to other tasks (e.g., object detection) and objects (e.g., roads, vehicles, vegetation, etc.).

We also explore several basic questions related to overhead SI: what is the impact of the quantity of synthetic imagery on performance, ablation studies of the city styles and training procedures, and an initial investigation into the mechanism by which Synthinel-1 is beneficial (e.g., matching the visual features of real worlds, or instilling robustness e.g., domain randomization [12], [13]). We will release the Synthinel-1 dataset with the publication of this work[1].

To our knowledge, we are the first to produce any of the important aforementioned results for overhead synthetic imagery. This work thereby provides researchers with the first well-validated baseline process for generating *useful* overhead imagery: a process that requires numerous steps and non-obvious design choices. This establishes an important foundation on which many additional lines of future work can be built (see Section 7).

## 2. Related Work

**Remote Sensing Datasets.** To support algorithm development, several publicly-available benchmark datasets have been developed for both segmentation, and general object recognition, on remote sensing imagery. A variety of objects have been considered for recognition in remote sensing imagery, such as buildings[5]–[7], roads[5], [15], [16], vehicles [17], solar arrays [18], and more [17], [19]. The most recent and most diverse datasets encompass hundreds of square kilometers of labeled imagery, collected over a few of distinct geographic locations. Some examples include DeepGlobe [5], Inria [4], DSTL[6], and ISPRS challenge[17]. Within a given geographic location, imagery is usually collected under similar conditions: e.g., the same sensor, day of year, and environmental conditions. Although these datasets are crucial to the current success of deep learning, they capture relatively little of the variability in real-world imagery, motivating the development of methods for generating synthetic imagery.

---

[1] https://github.com/timqqt/Synthinel

**Semantic Segmentation.** State-of-the-art segmentation algorithms are largely comprised of CNNs, and can be divided into two categories. The first category uses an encoder-decoder structure to maintain fine-grinned object boundary details [20], [21]. Variants of the U-net model, an encoder-decoder structure, recently yielded top performance on the Inria, DeepGlobe, and DSTL benchmarks for building segmentation in overhead imagery[6], [7], [22]. The second category makes use of feature pyramid pooling structures to capture contextual information at different image resolutions [23]–[25]. Variants of the DeepLab model, an example of this architecture, have recently led benchmarks for street view segmentation [26] and the PASCAL VOC segmentation challenge[27].

**Synthetic imagery for training networks.** In recent years interest has grown rapidly in the use of virtual worlds to generate ground truth, especially for semantic segmentation tasks where obtaining pixel-wise labels is especially time-consuming and costly. A large number of publications have demonstrated success using synthetic imagery [9]–[14], [28]. Some notable examples were presented in [9] (the SYNTHIA dataset) and [10], in which synthetic imagery with pixel-wise semantic labels were generated from 3D virtual worlds. It has shown that these synthetic datasets can boost the performance of segmentation networks on real-world benchmark imagery, such as for this task, such as the CamVid [29] and KITTI [30] dataset, among others. In [31] the authors were able to outperform standard training on real imagery using synthetic imagery that was stylized to look more realistic, using an adversarial loss, bridging the gap between synthetic and real imagery.

**Limitations of existing synthetic imagery rendering.** Several different resources have been developed for developing synthetic imagery such as those based on Unreal Engine [32], existing video game engines [28], the Unity game engine [9], [33]. Many of these engines can generate virtual worlds with high fidelity, but often rely on highly-designed models of objects, or layouts [12], [13] that can be time-consuming to construct. The existing models and tools for randomization are best designed for (relatively) small-scale scenes compared to overhead imagery, such as indoor scenes or street scenes. Designing objects like buildings, road networks, and vegetation for the large areas necessary for synthetic *overhead* imagery would be time-consuming, or infeasible. We explore CityEngine as a tool that can help fulfill these unique needs, providing rapid generation of randomized large-scale virtual worlds. We note however that CityEngine can be paired with many existing tools such as Unity, to improve visual fidelity where it may be helpful.

**Synthetic overhead imagery.** To our knowledge no rigorous work has been conducted on utilizing synthetic *overhead* imagery (e.g., satellite or high-altitude aerial photography) for training machine-learning models. The only existing work exploring this idea was recently (2018 presented in [34], however this work suffers from numerous limitations. Most crucially, the authors employed a military-grade rendering software that is inaccessible to the public, and their raw synthetic imagery was not made publicly available. This system also requires substantial design work to generate even very limited quantities of imagery. The authors only use a single private dataset for experimentation, rather than any benchmark datasets.

This contrasts with the systematic and comprehensive results provided here, as detailed in Section 1.2. Therefore, the work here essentially provides the first treatment of this topic in the research literature.

## 3. The Synthinel-1 dataset

In this section we briefly describes the process of creating Synthinel-1, as well as its characteristics, and an analysis of the costs/time associated with generating real-world and synthetic overhead imagery, respectively.

### 3.1. Synthetic imagery creation

We provide a brief overview here, but further details are available in the supplementary materials. Our methodology for generating synthetic imagery is built upon the CityEngine software[2]. Our main motive for using this software is that it allows users to *rapidly* generate geographically large virtual worlds, that are randomized to introduce variability, while being constrained to exhibit realistic characteristics (e.g., layouts of building, roads and landscapes; colors and textures of large-scale objects). Furthermore, the user is provided with high-level controls over the features of the virtual world.

For example, the software begins with a procedural generation algorithm for roadways (described in [35]). The designer can control features of the topology of the road network (e.g., "organic", "raster", "radial", or combinations). Once a street network is generated, the intervening space is randomly populated with structures (e.g., buildings, trees, landscapes, etc.). The designer can control the qualities of these objects using a combination of (i) libraries of object models and textures (easily customized if desired), and (ii) computer-generated Architecture (CGA) scripts (described in [36])**.**

We altered the CGA files to generate virtual worlds with desired geographic extent. We then developed Python scripts that communicate through a built-in CityEngine API to systematically move a simulated camera around the virtual world and take overhead photographs at regular spatial intervals. Our software also controls the camera height and field-of-view to obtain the proper resolution (approx. 0.3m/pixel).

---
[2] https://www.esri.com/en-us/arcgis/products/esri-cityengine/overview

### 3.2. Synthinel-1 dataset details

Although it is possible to design a virtual world with specific characteristics (e.g., to mimic a real-world city, or style), we began by leveraging pre-defined city "styles" that are freely available online. This is simpler, and we reasoned (if it worked) it would serve as an excellent baseline for more sophisticated approaches. We identified nine candidate styles to explore first, based on our subjective assessment of their realism. These styles are presented in Fig. 2, and comprise the full Synthinel-1 dataset.

We extracted 2,108 synthetic images with corresponding ground truth imagery, using the procedure described in Section 3.1. Each synthetic image is $572 \times 572$ pixels in size, with a resolution 0.3m/pixel. Our total pool of synthetic imagery is constructed from equal quantities of image patches from each of the nine styles illustrated in Fig. 2. We will release Synthinel-1 with the publication of this work.

*The Synth-1 subset*. For most experiments, unless otherwise stated, we use a subset of the full Synthinel-1 imagery composed of the following six styles: {a, b, c, g, h, i}. We refer to this as *Synth-1*. These styles were chosen based upon ablation studies in Section 6.1. The total number of images in Syn-1 is 1,640, or approximately 47 $km^2$ of labeled synthetic imagery.

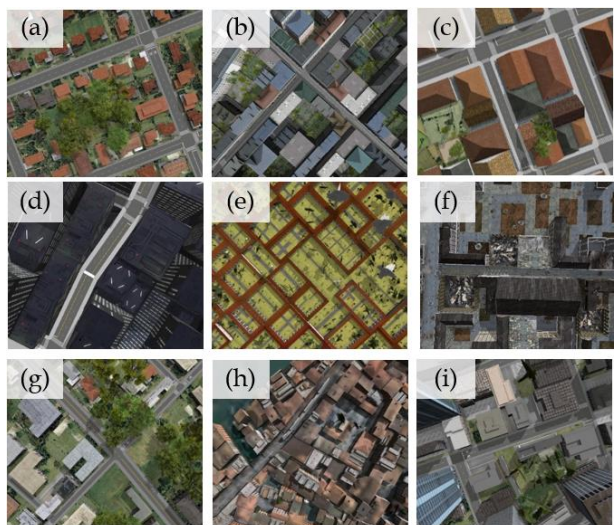

Fig. 2. An illustration of the nine different virtual city styles that we use. (a) Red roof style; (b) Paris' buildings style; (c) ancient building style; (d) sci-fi city style; (e) Chinese palace style;(f) Damaged city style (g) Austin city style; (h) Venice style; (i) modern city style.

### 3.3. Costs and time: real-world versus synthetic

In this section we compare real and synthetic imagery on two characteristics: the price of (i) acquiring imagery and (ii) annotation, respectively; comparing, for real and synthetic satellite imagery, respectively. The cost of satellite imagery can vary substantial, depending upon many factors: spatial resolution, geographic coverage, its age, level of preprocessing by the vendor, and more. However, archived (i.e., ≥3 month old) color imagery at 0.3m/pixel (i.e., a popular resolution, used in this work) from an imagery vendor encompassing a few municipal regions can cost on the order of tens of thousands of dollars, and in the millions for entire countries. By contrast, the marginal price for one additional square km of synthetic imagery is $0, once a commercial license to CityEngine is purchased at $2000 per year. Many research institutions (e.g., universities) also have a site license, making it free to use for students and researchers there. Therefore, even small quantities of real imagery greatly exceed the prices of synthetic imagery (using the approaches proposed here).

Annotation is another major cost (time and money) associated with using real overhead imagery. While synthetic imagery does not require any annotation – a major benefit – it still requires a software a designer to invest time designing features of the virtual world. As a result, a precise comparison of the time associated with each approach is difficult to make. For example, a carefully designed virtual world, with highly customized features intended to maximize realism, will require substantially greater design time than one that largely utilizes default settings. However, we hypothesize that designing a virtual world will generally require substantially less time than annotation, and we expect the advantage of synthetic imagery to grow as further research is conducted on its design. Finally, we note that, once a virtual world is designed, it requires (approx.) 1 minute per square km of generated overhead synthetic imagery on standard hardware (an Intel(R) Core(TM) i7-7700HQ CPU@2.80 GHz)  - negligible for most applications.

## 4. Experimental design details

In this section we first describe our default experimental design details. Particular details may vary in some experiments, and we will state this clearly where it occurs.

### 4.1. Satellite imagery benchmark datasets

**Inria.** The INRIA Aerial Image Labeling Challenge Dataset [7] is a popular recent benchmark dataset for building footprint segmentation. This dataset features RGB aerial imagery collected over ten cities across the U.S. and Europe. A total of 36 images were captured over each city, at a resolution of 0.3m. Each of the 36 images encompasses 2.25 $km^2$ resulting in 81$km^2$ of labeled imagery for each city. The ground truth for five of the cities is used as a benchmark performance metric (e.g., see [4]) and therefore it is not public. Therefore we conduct our experiments on the remaining five cities with publicly-available labels: Austin, Chicago, Kitsap, Western Tyrol, and Vienna.

**DeepGlobe (DG).** The DG dataset [5] is another popular benchmark recently utilized for a road and building footprint segmentation competition at CVPR 2018 [5]. The DG dataset features varying quantities of 0.3m imagery collected over four cities across the world: Shanghai, China ($133 \text{ km}^2$); Khartoum, Sudan ($29 \text{ km}^2$); Las Vegas, U.S.A. ($113 \text{km}^2$); and Paris, France ($33 \text{km}^2$).

### 4.2. Benchmark segmentation networks

For our experiments we consider two segmentation network architectures: U-net and DeepLabV3. There are (at least) two general architectures for segmentation: the encoder-decoder structure, and the feature pyramid structure. U-net and DeepLabV3 represent a popular version of each network architecture. The **U-net** was originally proposed for medical image segmentation [21], and has since become popular for segmentation of remote sensing imagery as well [4]. We use a modified version of the U-net model and training procedure that recently achieved the highest accuracy on the Inria benchmark competition [4], [37]. The **DeepLabV3** model, and its variants, have recently achieved state-of-the-art performance on the segmentation of street view scenes [26].

### 4.3. Network training details

Our networks are implemented in TensorFlow using the Adam optimizer to minimize a cross-entropy loss between the pixel-wise ground truth and predictions in each input patch. We train all networks for 80,000 mini-batch iterations with a batch size of seven. We found a batch size of seven to yield the best performance on real-world imagery, for both the U-net and DeepLabV3 models.

**Without synthetic imagery:** We use learning rates of 5e-5 and 1e-4 for the DeepLabV3 and U-net models, respectively. In contrast to DeepLabV3, the U-net does not have a pre-trained encoder, and therefore we found it performed best (training solely on real imagery) using a higher learning rate. For both networks we drop the learning rate by one order of magnitude after 50,000 iterations of training.

**With synthetic imagery added:** We employ a two-stage training procedure when using synthetic imagery. First we train using the mixed-batch training procedure previously employed for leveraging synthetic imagery in [10]. We again use a batch size of seven, but each batch contains six real images, and one synthetic image. In a second stage the model is fine-tuned on only the available real imagery for an additional 50,000 iterations using a reduced learning rate of 2e-5.

## 5. Benchmark testing with Synthinel-1

For these experiments we use the datasets, models, and training procedures described in Section 4. In these experiments we aim to evaluate two qualities of the synthetic imagery: within-sample and out-of-sample testing. The details and motives for these two data handling schemes is presented next (Section 5.1).

### 5.1. Data handling and performance metrics

We split the Inria and DG datasets into two disjoint subsets, as illustrated in Fig. 3. Following the guidance from the Inria authors[7], we use the first 5 tiles of each Inria city for testing (14%), and the remaining imagery for training (86%). For DG we used 60% of the imagery for training, and the remaining imagery for testing. All performance measures throughout the paper are computed on the same two testing subsets of the DG and Inria imagery. Synthetic imagery is only included for training models, in conjunction with real imagery (as opposed to testing). We use the intersection-over-union (IoU) metric for evaluating the performance of all trained models, following recent building segmentation studies [4], [5].

We have two goals with our data handling scheme. The first is to maintain constant testing datasets, so that we can isolate the impact of changes in the training strategy, and especially the inclusion of synthetic imagery. The second goal is to understand the impact of synthetic imagery when the trained model is evaluated on a novel imagery domain (i.e., imagery collected under novel imaging conditions, or at a new geographic location) with respect to the training imagery, versus a similar domain. Within-domain testing has historically been popular in the literature [6], [17], [38], but recent results [7], [8] indicate that the accuracy of deep learning models drops substantially when applied to novel data - a more challenging scenario, but arguably much more important for real-world application.

We propose to address these questions by training every model on just one of our two available real-world benchmark datasets, but evaluating them on both. When training with Inria, we use 14% of the training imagery (5 tiles) for validation. When training with DG, we use 10% of the training imagery for validation.

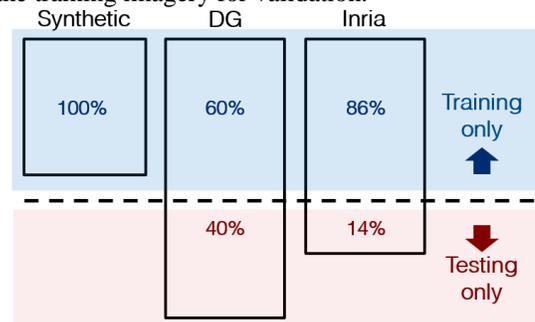

Fig. 3. Illustration of data handling for all experiments.

### 5.2. Testing on the training cities (within-domain)

The results of within-domain testing are presented in Table 1. In this case Synth-1 was beneficial in three of the four experiments with an average +0.4% improvement.

These results are unsurprising, since the models were tested on highly similar imagery – the testing tiles were sampled randomly from the total available imagery. These results are notable however because they indicate that Synth-1 is not detrimental even when the target domain is already highly similar to the training imagery – so in practice there is little risk to using Synth-1.

Table 1: Training and testing on *the same* geographic locations. Results (intersection-over-union) of segmentation on popular building segmentation benchmark datasets.

| Model | Scenario (train->test) | Add Synth-1? | IoU | % IoU Change |
|---|---|---|---|---|
| U-net | DG→DG | Yes | 0.682 | -0.6% |
| U-net | DG→DG | No | **0.686** | |
| U-net | Inria→Inria | Yes | **0.692** | 0.3% |
| U-net | Inria→Inria | No | 0.690 | |
| DeepLab V3 | DG→DG | Yes | **0.767** | 0.6% |
| DeepLab V3 | DG→DG | No | 0.762 | |
| DeepLab V3 | Inria→Inria | Yes | **0.730** | 1.1% |
| DeepLab V3 | Inria→Inria | No | 0.722 | |

### 5.3. Testing on previously unseen cities (out-of-domain)

The results of out-of-domain testing are presented in Table 2, indicating that the addition of Synth-1 always improves performance. The improvements are more substantial, ranging from 4.0% to nearly 20%, with an average of 9%. These results suggest that Synth-1 aids the models with generalization to novel imagery, collected under different conditions, or in different locations.

Table 2: Training and testing on *different* geographic locations. Results (intersection-over-union) of segmentation on popular building segmentation benchmark datasets.

| Model | Scenario (train->test) | Add Synth-1? | IoU | % IoU Change |
|---|---|---|---|---|
| U-net | DG→Inria | Yes | **0.529** | 6.8% |
| U-net | DG→Inria | No | 0.495 | |
| U-net | Inria→DG | Yes | **0.247** | 19.9% |
| U-net | Inria→DG | No | 0.206 | |
| DeepLab V3 | DG→Inria | Yes | **0.624** | 4.0% |
| DeepLab V3 | DG→Inria | No | 0.600 | |
| DeepLab V3 | Inria→DG | Yes | **0.404** | 6.6% |
| DeepLab V3 | Inria→DG | No | 0.379 | |

These results are especially compelling because Synth-1 is raw synthetic imagery – recent techniques for domain adaptation and style transfer have been used to improve the utility of synthetic imagery, and could likely be adapted here to substantially improve the value of the synthetic imagery. Furthermore, a more systematic exploration of styles and imaging conditions of the synthetic imagery (e.g., lighting angles, intensity, and camera angles) could yield further improvements as well.

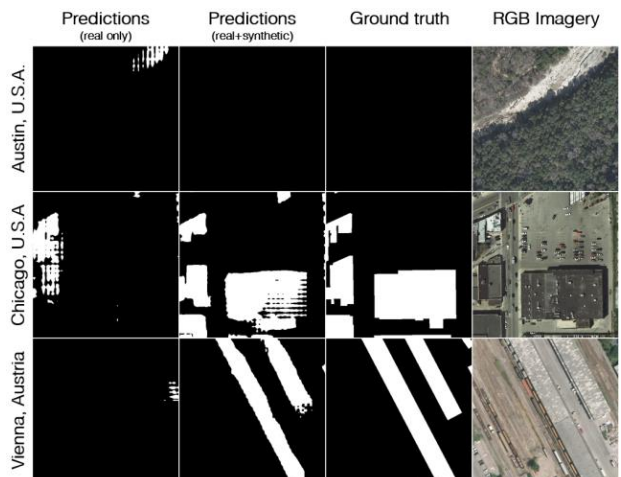

Fig. 4. Examples of predictions made by the DeepLabV3 model before, and after, the inclusion of synthetic imagery in training. Three examples are shown along the rows, illustrating diverse geographic locations.

Comparing the results in Table 2 and Table 1, we also see that there is a substantial performance loss when models are tested on a new domain, corroborating recent evidence [7], [8] that this is a problem. The results here indicate that synthetic imagery may be a viable avenue to help overcome this practical challenge, acting as a complement to other techniques for visual domain adaptation [11], [39].

In Fig. 4 we presents examples of predictions made by DeepLabV3 for the DG → Inria scenario (Table 5), providing qualitative examples of cases when the benefits of the synthetic imagery.

### 5.4. Why is Synthinel-1 helpful: domain matching, or domain confusion?

Here we stratify the results in Section 5.3 by city, and investigate whether the addition of Synth-1 is beneficial for performance on particular cities, or it tends to improve performance across all real-world cities. The results for DG and Inria are presented in Tables 3 and 4, respectively. The results indicate that, while there are variations in the degree of performance improvement across cities, there appears to be no strong bias in favor of one city. We hypothesize therefore that the benefits of Synth-1 are most similar to those of domain randomization[12], [13], in which models are improved by presenting them with synthetic data exhibiting diverse and possibly unrealistic visual features.

Table 3: City-wise performance (IoU) when evaluating DeepLabV3 in the DG→Inria scenario.

| Training data | DeepGlobe testing city | | | |
|---|---|---|---|---|
| | Vegas | Shanghai | Paris | Khartoum |
| DG+Synth-1 | **0.633** | **0.365** | **0.477** | **0.268** |
| DG | 0.598 | 0.155 | 0.396 | 0.072 |

In contrast, our virtual cities could provide the model with styles of cities that are present in some of the benchmark data – an effect we term as domain matching.

This would allow the model to learn the particular textures and colors of the buildings in the particular target cities of interest, but would not necessarily lead to a more robust model overall. In this case we would expect to see strong performance improvements on particular cities, rather than a consistent improvement across all real cities.

Table 4: City-wise performance (IoU) when evaluating DeepLabV3 in the Inria→DG scenario.

| Training data | Inria testing city | | | | |
|---|---|---|---|---|---|
| | Austin | Chicago | Kitsap | Tyrol-w | Vienna |
| Inria+Synth-1 | **0.602** | **0.580** | **0.573** | **0.640** | **0.690** |
| Inria | 0.582 | 0.548 | 0.565 | 0.600 | 0.670 |

### 5.5. Blind testing on an additional benchmark

In an effort to further validate Synth-1, we evaluated the models that were trained on DG and Inria in Section 5.3 on the ISPRS benchmark (2016) [17] for multiclass segmentation. We used the models, *as is*, with no optimization, and applied blindly to ISPRS. In the ISPRS dataset we treated all ground truth object classes as a single background class, except for the building class. We also resampled the imagery to match the resolution of DG and Inria. The results are presented in Table 5, and indicate that Synth-1 provides large performance improvements three of the four cases. In the one case it failed (U-net trained on DG), the performance of the U-net was already extremely low (IoU=0.15), suggesting that the models are making highly random predictions, and the performance variations across the two models may be dominated by factors other than their ability to recognize buildings (e.g., a larger/smaller prior of predictions in favor of one class).

Table 5: Results (intersection-over-union) of blind segmentation of buildings on the ISPRS benchmark.

| Model | Scenario (train->test) | Add Synth-1? | IoU | % IoU Change |
|---|---|---|---|---|
| U-net | DG→ISPRS | Yes | **0.133** | -13.6% |
| | DG→ISPRS | No | 0.154 | |
| | Inria→ISPRS | Yes | **0.477** | +6% |
| | Inria→ISPRS | No | 0.450 | |
| DeepLab V3 | DG→ISPRS | Yes | **0.683** | +1.3% |
| | DG→ISPRS | No | 0.674 | |
| | Inria→ ISPRS | Yes | **0.635** | +9.3% |
| | Inria→ISPRS | No | 0.581 | |

## 6. Additional analysis

In this section we provide some further analysis of the properties of overhead SI, using the Synth-1 dataset.

### 6.1. Ablation and training optimization studies

In this section we systematically vary different characteristics of the Synth-1 dataset, as well as our training procedure, and evaluate their impact. All of these experiments were conducted using the DeepLabV3 model with the DG → Inria scenario. Due to space limitations, we summarize the findings here, but the full experimental results can be found in the supplementary material:

- **City styles in Synth-1:** We could not exhaustively explore the impact of including/excluding all combinations of city styles in Synth-1, so we incrementally removed city styles and evaluated its impact on performance. Based on these experiments we found styles {a, b, c, g, h, i} yielded the best results.
- **Training with synthetic imagery:** We explored two recent strategies from the literature for training with synthetic imagery: Mixed-batch (MB) [10], Balanced Gradient Contribution (BCG) [9]. For MB we varied the mini-batch ratio and for BCG we varied the weights assigned to real and synthetic imagery, respectively. We found MB training with a batch ratio of 6:1 (real:synthetic) yielded the best results.
- **Fine-tuning on real data:** We considered fine-tuning on real data as a second stage of training, following joint training (i.e., mixed-batch training) using both synthetic and real imagery. We found fine-tuning was consistently beneficial.

Through this exploration we improved the performance of models using Synth-1, but we note that Synth-1 was usually beneficial, even without optimal settings.

### 6.2. The impact of the quantity of synthetic imagery

In this section we varied the quantity of either real and SI, separately, and evaluated its impact on the performance of the models. Once again we use the DeepLabV3 model and the DG→Inria scenario. In Fig. 5 we gradually reduce the size of the Synth-1 dataset by randomly sampling and removing tiles from it.

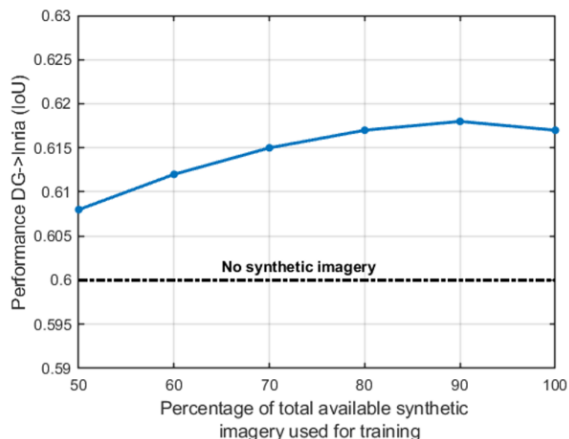

Fig. 5. Performance (IoU) versus the quantity of unique synthetic imagery available.

The results indicate that performance saturates at a size of 70-80% of the Synth-1 size. This suggests that, given our current approach for randomly generating imagery, there is little additional benefit beyond roughly 6 $km^2$ of imagery (~80% total available) from each style of virtual city. Introduction of further sources of randomness (e.g., additional building shapes, sizes, background scenery) could however result in the need for larger areas of imagery for each individual style.

## 7. Conclusions and future work

**Conclusions.** In this work we explored the use of synthetic *overhead* imagery for training deep learning models for segmentation in overhead imagery. We developed software tools for rapidly generating synthetic overhead imagery, and used the tools to generate a set of overhead imagery, termed Synthinel-1, that we release with this publication. We further demonstrated that Synth-1 (a subset of Synthinel-1) can be used to augment real satellite imagery to improve the performance of building segmentation models, especially on novel imagery that was not present in the training dataset.

To our knowledge, we are the first to produce any of the aforementioned results for *overhead* synthetic imagery. This work thereby provides researchers with the first well-validated baseline process for generating *useful* overhead imagery: a process that requires numerous steps and non-obvious design choices. This establishes an important foundation on which many additional lines of future work can be built (see Section 7).

**Potential future work.** There are many potential avenues to improve and expand upon the work presented here. We began here with building segmentation due to its popularity, but these approaches can easily be extended to other tasks (e.g., object detection) and objects (e.g., roads, vehicles, vegetation, etc.). Another important avenue of exploration is the introduction of variability in the lighting conditions, camera angle, image resolution, and other factors that are extremely difficult to obtain in real-world imagery, but easily be introduced with synthetic imagery. We did not provide a systematic investigation of these factors here. Another important avenue of subsequent work is applying more recent and sophisticated forms of domain adaptation (e.g., [11]) to further improve the utility of Synth-1.

## Acknowledgments

We thank the NVIDIA Corporation, the NSF XSEDE computational environment, and Duke Research Computing for providing computing resources for this work. We also thank the Duke University Energy Initiative for their support for this work. We also thank the reviewers for their time and feedback, which greatly improved this work.

## References

[1] J. A. Benediktsson and M. Pesaresi, "A new approach for the morphological segmentation of high-resolution satellite imagery," *IEEE Trans. Geosci. Remote Sens.*, vol. 39, no. 2, pp. 309–320, 2001.

[2] X. Jin and C. H. Davis, "Automated Building Extraction from High-Resolution Satellite Imagery in Urban Areas Using Structural, Contextual, and Spectral Information," *EURASIP J. Adv. Signal Process.*, vol. 2005, no. 14, pp. 2196–2206, 2005.

[3] B. Sirmacek and C. Unsalan, "Urban-area and building detection using SIFT keypoints and graph theory," *IEEE Trans. Geosci. Remote Sens.*, vol. 47, no. 4, pp. 1156–1167, 2009.

[4] B. Huang *et al.*, "Large-scale semantic classification: outcome of the first year of inria aerial image labeling benchmark," in *International Geoscience and Remote Sensing Symposium*, 2018.

[5] I. Demir *et al.*, "DeepGlobe 2018: A Challenge to Parse the Earth through Satellite Images," pp. 172–181, 2018.

[6] V. Iglovikov, S. Mushinskiy, and V. Osin, "Satellite Imagery Feature Detection using Deep Convolutional Neural Network: A Kaggle Competition," vol. June, 2017.

[7] E. Maggiori *et al.*, "Can Semantic Labeling Methods Generalize to Any City ? The Inria Aerial Image Labeling Benchmark To cite this version :," pp. 3226–3229, 2017.

[8] R. Wang *et al.*, "The poor generalization of deep convolutional neural networks to aerial imagery from new geographic locations: an empirical study with solar array detection," in *IEEE Applied Imagery Pattern Recognition Workshop*, 2017, pp. 1–9.

[9] G. Ros, L. Sellart, J. Materzynska, D. Vazquez, and A. M. Lopez, "The SYNTHIA Dataset: A Large Collection of Synthetic Images for Semantic Segmentation of Urban Scenes," *Proc. IEEE Comput. Soc. Conf. Comput. Vis. Pattern Recognit.*, vol. 2016-Decem, no. 600388, pp. 3234–3243, 2016.

[10] S. R. Richter, V. Vineet, S. Roth, and V. Koltun, "Playing for Data: Ground Truth from Computer Games," in *European conference on computer vision*, 2016, vol. 9908, pp. 102–118.

[11] S. Sankaranarayanan, Y. Balaji, A. Jain, S. N. Lim, and R. Chellappa, "Learning from Synthetic Data: Addressing Domain Shift for Semantic Segmentation," in *Computer Vision and Pattern Recognition*, 2018.

[12] J. Tremblay *et al.*, "Training deep networks with synthetic data: Bridging the reality gap by domain randomization," *IEEE Comput. Soc. Conf. Comput. Vis. Pattern Recognit. Work.*, vol. 2018-June, pp. 1082–1090, 2018.

[13] J. Tobin, R. Fong, A. Ray, J. Schneider, W. Zaremba, and P. Abbeel, "Domain randomization

for transferring deep neural networks from simulation to the real world," *IEEE Int. Conf. Intell. Robot. Syst.*, vol. 2017-Septe, pp. 23–30, 2017.

[14] Y. Zhang *et al.*, "Physically-based rendering for indoor scene understanding using convolutional neural networks," *Proc. - 30th IEEE Conf. Comput. Vis. Pattern Recognition, CVPR 2017*, vol. 2017-January, pp. 5057–5065, 2017.

[15] K. Bradbury, B. Brigman, L. M. Collins, T. Johnson, S. Lin, and R. Newell, "Arlington, Massachusetts - Aerial imagery object identification dataset for building and road detection, and building height estimation," 2016. [Online]. Available: https://figshare.com/articles/Arlington_Massachusetts_-_Aerial_imagery_object_identification_dataset_for_building_and_road_detection_and_building_height_estimation/3485204.

[16] V. Mnih and G. E. Hinton, "Learning to detect roads in high-resolution aerial images," *Lect. Notes Comput. Sci. (including Subser. Lect. Notes Artif. Intell. Lect. Notes Bioinformatics)*, vol. 6316 LNCS, no. PART 6, pp. 210–223, 2010.

[17] S. Paisitkriangkrai, J. Sherrah, P. Janney, and A. Van Den Hengel, "Semantic Labeling of Aerial and Satellite Imagery," *IEEE J. Sel. Top. Appl. Earth Obs. Remote Sens.*, vol. 9, no. 7, pp. 2868–2881, 2016.

[18] K. Bradbury *et al.*, "Distributed Solar Photovoltaic Array Location and Extent Data Set for Remote Sensing Object Identification," *figshare*, 2016. [Online]. Available: http://www.nature.com/articles/sdata2016106. [Accessed: 01-Jun-2016].

[19] G.-S. Xia *et al.*, "DOTA: A large-scale dataset for object detection in aerial images," in *Proceedings of the IEEE Conference on Computer Vision and Pattern Recognition*, 2018, pp. 3974–3983.

[20] V. Badrinarayanan, A. Kendall, and R. Cipolla, "Segnet: A deep convolutional encoder-decoder architecture for image segmentation," *IEEE Trans. Pattern Anal. Mach. Intell.*, vol. 39, no. 12, pp. 2481–2495, 2017.

[21] O. Ronneberger, P. Fischer, and T. Brox, "U-Net: Convolutional Networks for Biomedical Image Segmentation," *Miccai*, pp. 234–241, 2015.

[22] V. I. Iglovikov, S. Seferbekov, A. V. Buslaev, and A. Shvets, "TernausNetV2: Fully Convolutional Network for Instance Segmentation," 2018.

[23] L.-C. Chen, G. Papandreou, F. Schroff, and H. Adam, "Rethinking Atrous Convolution for Semantic Image Segmentation," 2017.

[24] H. Zhao, J. Shi, X. Qi, X. Wang, and J. Jia, "Pyramid scene parsing network," *Proc. - 30th IEEE Conf. Comput. Vis. Pattern Recognition, CVPR 2017*, vol. 2017-Janua, pp. 6230–6239, 2017.

[25] T.-Y. Lin, P. Dollár, R. Girshick, K. He, B. Hariharan, and S. Belongie, "Feature Pyramid Networks for Object Detection," *Proc. IEEE Conf. Comput. Vis. Pattern Recognit.*, pp. 2117–2125, 2016.

[26] L. Chen, Y. Zhu, G. Papandreou, F. Schroff, and C. V Aug, "Encoder-Decoder with Atrous Separable Convolution for Semantic Image Segmentation," 2018.

[27] M. Everingham, L. Van Gool, C. K. I. Williams, J. Winn, and A. Zisserman, "The pascal visual object classes (VOC) challenge," *Int. J. Comput. Vis.*, vol. 88, no. 2, pp. 303–338, 2010.

[28] A. Shafaei, J. J. Little, and M. Schmidt, "Play and learn: Using video games to train computer vision models," *Br. Mach. Vis. Conf. 2016, BMVC 2016*, vol. 2016-September, pp. 26.1-26.13, 2016.

[29] G. J. Brostow, J. Fauqueur, and R. Cipolla, "Semantic object classes in video: A high-definition ground truth database," *Pattern Recognit. Lett.*, vol. 30, no. 2, pp. 88–97, 2009.

[30] A. Geiger, P. Lenz, C. Stiller, and R. Urtasun, "Vision meets robotics: The KITTI dataset," *Int. J. Rob. Res.*, vol. 32, no. 11, pp. 1231–1237, 2013.

[31] A. Shrivastava, T. Pfister, O. Tuzel, J. Susskind, W. Wang, and R. Webb, "Learning from simulated and unsupervised images through adversarial training," *Proc. - 30th IEEE Conf. Comput. Vis. Pattern Recognition, CVPR 2017*, vol. 2017-Janua, pp. 2242–2251, 2017.

[32] W. Qiu and A. Yuille, "UnrealCV: Connecting Computer Vision to Unreal Engine," pp. 467–474, 2016.

[33] A. Gaidon, Q. Wang, Y. Cabon, and E. Vig, "VirtualWorlds as Proxy for Multi-object Tracking Analysis," *Proc. IEEE Comput. Soc. Conf. Comput. Vis. Pattern Recognit.*, vol. 2016-December, pp. 4340–4349, 2016.

[34] S. Han *et al.*, "Efficient generation of image chips for training deep learning algorithms," vol. 1020203, no. May 2017, p. 1020203, 2017.

[35] G. Chen, G. Esch, P. Wonka, P. Müller, and E. Zhang, "Interactive procedural street modeling," in *SIGGRAPH*, 2008, vol. 27, no. 3, p. 1.

[36] P. Muller, P. Wonka, S. Haegler, A. Ulmer, and L. Van Gool, "Procedural modeling of buildings," in *SIGGRAPH*, 2006, vol. 1, no. 212, pp. 1–10.

[37] B. Huang, D. Reichman, L. M. Collins, K. Bradbury, and J. M. Malof, "Sampling training images from a uniform grid improves the performance and learning speed of deep convolutional segmentation networks on large

aerial imagery," in *IGARSS*, 2018.

[38] S. Golovanov, R. Kurbanov, A. Artamonov, A. Davydow, and S. Nikolenko, "Building detection from satellite imagery using a composite loss function," *IEEE Comput. Soc. Conf. Comput. Vis. Pattern Recognit. Work.*, vol. 2018-June, pp. 219–222, 2018.

[39] V. M. Patel, R. Gopalan, R. Li, and R. Chellappa, "Visual Domain Adaptation: A survey of recent advances," *IEEE Signal Process. Mag.*, vol. 32, no. 3, pp. 53–69, 2015.